\documentclass{article}

\usepackage{arxiv}

\usepackage[utf8]{inputenc} 
\usepackage[T1]{fontenc}    
\usepackage{hyperref}       
\usepackage{url}            
\usepackage{booktabs}       
\usepackage{amsfonts}       
\usepackage{nicefrac}       
\usepackage{microtype}      
\usepackage{lipsum}
\usepackage{graphicx}
\usepackage{amsmath}
\graphicspath{ {./images/} }

\title{GS-KGC: A Generative Subgraph-based Framework for Knowledge Graph Completion with Large Language Models}

\author{
Rui Yang$^{1}$, Jiahao Zhu$^{2}$, Jianping Man$^{1}$, Hongze Liu$^{1}$, Li Fang$^{1,}$\footnote{Co-corresponding author. E-mail address: fangli@mail.sysu.edu.cn}, Yi Zhou$^{1,}$\footnote{Corresponding author. E-mail address: zhouyi@mail.sysu.edu.cn} \\
\\
$^1$Zhongshan School of Medicine, Sun Yat-sen University \\
$^2$School of Computer, Sun Yat-sen University \\
}

\begin{document}
\maketitle
\begin{abstract}
Knowledge graph completion (KGC) focuses on identifying missing triples in a knowledge graph (KG) , which is crucial  for many downstream applications. Given the rapid development of large language models (LLMs), some LLM-based methods are proposed for KGC task. However, most of them focus on prompt engineering while overlooking the fact that finer-grained subgraph information can aid LLMs in generating more accurate answers. In this paper, we propose a novel completion framework called \textbf{G}enerative \textbf{S}ubgraph-based KGC (GS-KGC), which utilizes subgraph information as contextual reasoning and employs a QA approach to achieve the KGC task. This framework primarily includes a subgraph partitioning algorithm designed to generate negatives and neighbors. Specifically, negatives can encourage LLMs to generate a broader range of answers, while neighbors provide additional contextual insights for LLM reasoning. Furthermore, we found that GS-KGC can discover potential triples within the KGs and new facts beyond the KGs. Experiments conducted on four common KGC datasets highlight the advantages of the proposed GS-KGC, e.g., it shows a 5.6\% increase in Hits@3 compared to the LLM-based model CP-KGC on the FB15k-237N, and a 9.3\% increase over the LLM-based model TECHS on the ICEWS14.
\end{abstract}


\section{Introduction}
\label{sec:introduction}
A knowledge graph (KG) is a structured semantic knowledge base that organizes entities, concepts, attributes, and their relationships in a graph format. In a KG, nodes represent entities or concepts, and edges depict the relationships between them. For instance, a KG in the film domain might include entities like \textit{movie}, \textit{director}, and \textit{actor}, along with relationships such as \textit{directed} and \textit{acted in}. This structure facilitates complex queries and reasoning tasks, proving valuable in areas like semantic search, recommendation systems, and natural language processing (NLP). However, due to limited annotation resources and technical constraints, existing KGs often have missing key entities or relationships, limiting their functionality in downstream tasks. To address this, knowledge graph completion (KGC) has been developed to infer, predict, and fill in these missing entities and relationships by analyzing existing triples, thereby enhancing the value and effectiveness of KGs in practical applications.

Previous research has employed various methods to address the problem of KGC. Among these, KG embedding methods like TransE\cite{bordes2013translating} and ComplEx\cite{trouillon2016complex} map entities and relationships into a low-dimensional vector space and use scoring functions to evaluate triples for reasoning. These methods primarily focus on the structural information of KGs, often neglecting the rich semantic content of entities. To address this limitation, text-based KGC models like KG-BERT\cite{yao2019kg} and SimKGC\cite{wang2022simkgc} have emerged, utilizing pre-trained language (PLM) models to capture detailed semantic information of entities. Language model and text-based methods have demonstrated strong performance on certain datasets. These methods often cannot directly produce the predicted results; they instead depend on ranking candidate entities, which confines their predictions to the current dataset. Recently, large language models (LLMs) have shown exceptional generative capabilities across various fields, inspiring researchers to explore their potential applications in KGC tasks.
\begin{figure}[h!]
    \centering
    \includegraphics[width=1.0\linewidth]{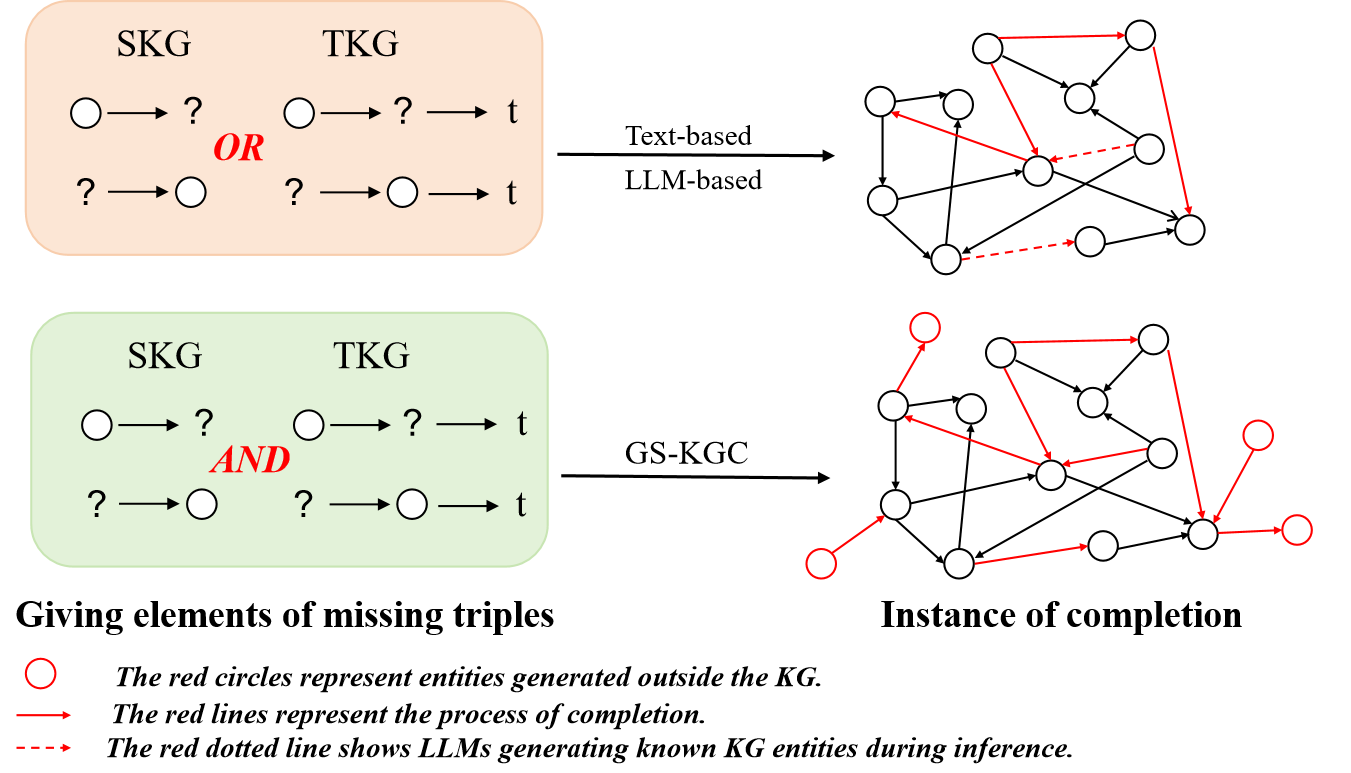} 
    \caption{Compared with previous KGC methods, generative KGC can discover new facts outside the KG.}
    \label{fig:problem}
\end{figure}

Current research in KGC typically involves separate methods for static knowledge graphs (SKGs) and temporal knowledge graphs (TKGs), focusing primarily on completing information within existing KGs, as shown in Figure 1. Methods based on LLMs have shown promising results but still have room for improvement in terms of performance. For instance, Yang et al. proposed CP-KGC\cite{yang2024enhancing}, which uses LLMs to enhance the semantic representation of text. However, this approach remains a text-based KGC strategy. Luo et al.\cite{luo2024chain} introduced COH, which uses historical information from TKGs to help LLMs infer recent events. While this method enables learning new patterns from history, it still has performance limitations. Similar to real-world events, missing triples in KGs often have multiple answers. LLMs struggle with accurate reasoning without effective context, limiting their predictive performance in closed datasets. Additionally, LLMs are trained on vast datasets, so their reasoning may align more with real-world facts than standardized answers. Research in this area is still relatively lacking.

To address these issues, we propose a novel completion framework called \textbf{G}enerative \textbf{S}ubgraph-based KGC (GS-KGC), which leverages LLMs to perform KGC by generating subgraph-based questions and deriving answers from them. Specifically, we design questions for incomplete triples needing prediction, such as $(h,r,?)$ or $(?,r,t)$, where $h+r$ or $t+r$ forms part of the question, and the predicted entity is the answer. During training, the $h+r$ or $t+r$ is divided into input and supervised annotation parts. In the generated questions, we propose a strategy that extracts subgraphs centered on entities and relationships within the KG. This strategy incorporates recalled negatives and neighbors to assist LLMs in adapting to specific events and complex logical relationships. To improve training efficiency, we filtered and pruned the recalled negative samples and neighborhood information.

Our contributions are as follows:

\begin{itemize}
    \item We propose a subgraph-based generative KGC framework called GS-KGC. This framework innovatively leverages negatives and neighbors to encourage LLMs to generate more potential answers, thereby improving the predictive ability of LLMs when dealing with situations involving multiple answers to a question. 
    \item GS-KGC can not only discover potential triples within the KG but also predict new entities outside of the graph, aiding in the discovery of new facts. The generative capabilities of LLMs endow GS-KGC with strong scalability and generalization capabilities. 
    \item Experiments on four common KGC datasets indicate the superiority of our method, which shows that our GS-KGC surpasses current text-based and LLM-based methods. Furthermore, the ablation study demonstrates the effectiveness of each module in the GS-KGC framework.
\end{itemize}

The remainder of this paper is organized as follows: Section \ref{sec:Related Work} introduces prior research pertinent to this study. Section \ref{sec:Methodlogy} describes the implementation of subgraph partitioning and the training of the GS-KGC framework. Section \ref{sec:Experiment} details the experimental results. Section \ref{sec:Analysis} analyzes the impact of hyperparameters and presents ablation studies, discussing the advantages of generative KGC. Section \ref{sec:Conclusion} presents a summary and considerations for future research.

\section{Related Work}\label{sec:Related Work}
\subsection{Knowledge Graph Completion}

KGC aims to fill in missing entities and links in a KG by learning and inferring stable relationships and attributes, thereby enhancing the graph's completeness. KGC is generally categorized into embedding-based and text-based approaches.

\textbf{Embedding-based KGC}: TransE\cite{bordes2013translating} is a pioneering embedding model that addresses link prediction by viewing relations as linear translations between entities. Subsequent models, TransH\cite{wang2014knowledge} and TransD\cite{ji2015knowledge}, improve upon TransE by introducing relation-specific hyperplanes and dynamic mapping matrices, which better handle complex relationships like one-to-many, many-to-one, and many-to-many. DistMult\cite{yang2014embedding} is a simple and efficient bilinear model that uses symmetric weight matrices to capture relationships between entities. However, it cannot handle the directionality of relationships. To address this issue, ComplEx\cite{trouillon2016complex} introduces complex-valued representations, enabling the model to distinguish antisymmetric relationships. RotatE\cite{sun2019rotate} models relationships by rotating entities in the complex space, capturing more complex relationship patterns. TuckER\cite{balavzevic2019tucker}, based on tensor decomposition techniques, uses a core tensor and two embedding matrices to predict entities and relationships effectively, demonstrating superior generalization capabilities. While these embedding models have made significant progress in KGC, they often fail to capture the high-level semantic features of entities and relationships, limiting their adaptability in more complex or dynamic environments.

\textbf{Text-based KGC}: Text-based KGC leverages PLMs to enhance the representation and understanding of entity relationships using detailed textual descriptions. For instance, KG-BERT by Yao et al.\cite{yao2019kg} uses BERT\cite{devlin2018bert} to encode triples and their textual descriptions into continuous text sequences, providing a foundational method for integrating deep contextual information into KGC. StAR by Wang et al.\cite{wang2021structure} builds on KG-BERT by combining graph embedding techniques with text encoding, using a Siamese-style encoder to enhance structured knowledge and achieve more nuanced entity representations. Despite these advancements, text-based methods generally lagged behind embedding-based approaches in performance until the introduction of SimKGC\cite{wang2022simkgc}. BERT-FKGC\cite{li2024bert} enhances few-shot KGC by incorporating textual descriptions, demonstrating the potential of pretrained language models in this area. SimKGC demonstrated, for the first time, superiority over embedding methods on certain datasets, proving the potential of integrating contrastive learning techniques with PLMs to significantly enhance performance.

\subsection{LLM-enhanced KGC}

\textbf{LLMs with Parameter-Efficient Fine-tuning:} Recent LLMs like Qwen\cite{yang2024qwen2}, Llama\cite{touvron2023llama}, and GPT-4\cite{achiam2023gpt} have gained attention for their exceptional emergent abilities\cite{zhao2023survey,wei2022emergent} and have significantly improved applications for enhancing KGs\cite{pan2024unifying}. These models can generate new facts for KG construction using simple prompts\cite{zhou2022large,shin2020autoprompt,brown2020language,jiang2020can}. Additionally, the introduction of chain-of-thought techniques has significantly enhanced the reasoning capabilities of LLMs. Wei et al.\cite{wei2022chain} introduced the chain-of-thought prompting method, which greatly improved the model's performance on arithmetic, common sense, and symbolic reasoning tasks by incorporating intermediate reasoning steps in the prompts.

Recent research has introduced various parameter-efficient fine-tuning techniques (PEFT) to optimize the performance of PLMs. One method involves adding adapters - small trainable feed forward networks - into the architecture of PLMs to enhance their functionality\cite{he2021towards,rebuffi2017learning,houlsby2019parameter,bapna2019simple}. Additionally, the Low-Rank Adaptation (LoRA) technique\cite{hu2021lora} simplifies model parameter updates by using low-dimensional representations during the fine-tuning process. Furthermore, prompt tuning\cite{lester2021power} and prefix tuning\cite{li2021prefix} enhance model input or activation functions by introducing learnable parameters, thus improving the model's adaptability and efficiency for specific tasks. These techniques improve the flexibility and applicability of models and enhance their ability to quickly adapt to new tasks while maintaining the stability of the original model structure.

\textbf{LLM-based KGC}: Since the introduction of the KICGPT framework in KGC research\cite{wei2024kicgpt}, KICGPT has successfully guided LLMs to generate rich contextual paragraphs using a triple-based KGC retriever, supporting small-scale KGC models and significantly enhancing their performance. Research shows that LLMs have substantial potential in understanding and generating rich semantic information. The CP-KGC\cite{yang2024enhancing} framework further develops this approach by enhancing text-based KGC methods through semantic enrichment and contextual constraints. This framework increases semantic richness and effectively identifies polysemous entities in KGC datasets by designing prompts tailored to different datasets. Additionally, the KG-LLM\cite{yao2023exploring} method treats triples as text sequences and adjusts LLMs with instructions to evaluate the reasonableness of triples or candidate entities and relationships, showing potential for high performance in triple classification and relationship prediction tasks. The MPIKGC\cite{xu-etal-2024-multi} framework enhances KGC from multiple angles by querying LLMs to expand entity descriptions, understand relationships, and extract structures. As research progresses, the implementation of LLMs in temporal KGC (TKGC) has been extensively studied, requiring the use of established temporal structures to predict missing event links at future timestamps. The application of LLMs in this field has improved reasoning ability on temporal structures through enhanced historical modeling and the introduction of inverse logic\cite{luo2024chain}. Additionally, Yuan et al.\cite{yuan2024back} studied that the performance of LLMs in complex temporal reasoning tasks can be significantly improved through high-quality instruction fine-tuning datasets. These approaches improve model performance in link prediction and triple classification tasks, demonstrating the potential for integrating various enhancement methods to improve overall performance.

This body of research creates a rich landscape of LLM applications in KGC, showcasing the broad possibilities and practical benefits of using LLMs to enhance KGC tasks from multiple perspectives. However, the performance of current LLM-based KGC research does not match that of mainstream models and fails to fully assess the potential impact of LLMs on KGC.

\section{Methodology}\label{sec:Methodlogy}
In this section, we introduce a novel generative KGC method named GS-KGC. As shown in Figure 2, it comprises three modules:

\begin{figure}[h!]
    \centering
    \includegraphics[width=1.0\linewidth]{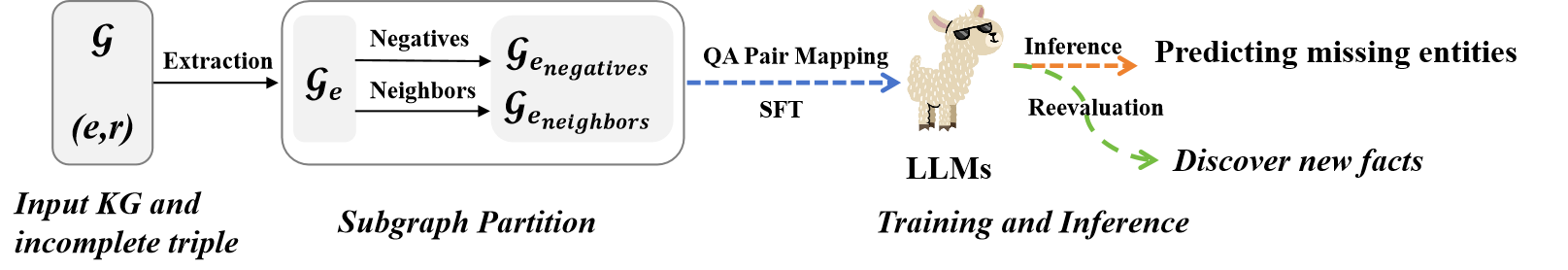} 
    \caption{GS-KGC model architecture.}
    \label{fig:method-GS}
\end{figure}

\textbf{Subgraph Partition}: This module divides the subgraph into two parts: a set of negatives and a set of neighbors, based on known entity and relation. This information provides context during model training.

\textbf{Training and inference}: This module transforms the traditional KGC task into a question-answering (QA) task, implementing link prediction through a fine-tuned LLM.

\textbf{Reevaluate new facts}: This module aids in reassessing the validity of new events predicted by link prediction task. It helps uncover new relationships within the KG and extends beyond the constraints of closed datasets.

\subsection{Subgraph partition}
A KG is defined as $\mathcal{G} = \{\mathcal{E}, \mathcal{R}, \mathcal{T}\}$. $\mathcal{E}$ is the entity set, which includes individuals such as persons, locations, and organizations. $\mathcal{R}$ is the relation set, encompassing relations between entities such as \textit{hasFriend} and \textit{locatedIn}. $\mathcal{T} = \{(h, r, t)|h \in \mathcal{E}, r \in \mathcal{R}, t \in \mathcal{E}\}$ is the triple set, where $h$, $r$, and $t$ represent the head entity, relation, and tail entity of the triple, respectively. Given $\mathcal{G}$, we can obtain $\mathcal{T}_{\text{train}}$, $\mathcal{T}_{\text{valida}}$, and $\mathcal{T}_{\text{test}}$.  When model predict entities, each triple must be predicted bidirectionally, considering forward prediction $(h, r, ?)$ and backward prediction $(?, r, t)$. We unify these two types of missing triples as $(e, r)$, where $e$ represents the existing head or tail entity, and $r$ represents the relation. According to the above definition, we obtain subgraphs for each triple separately and acquire the subgraph collection $\mathcal{G}_{\text{part}} = \{\mathcal{G}_1, \mathcal{G}_2, \ldots, \mathcal{G}_m\}$, $m = 2 \times (|\mathcal{T}_{\text{train}}| + |\mathcal{T}_{\text{valida}}| + |\mathcal{T}_{\text{test}}|)$. Notably, the subgraph data in $\mathcal{T}_{\text{valida}}$ and $\mathcal{T}_{\text{test}}$ are derived from $\mathcal{T}_{\text{train}}$.
\begin{figure}[h!]
    \centering
    \includegraphics[width=1.0\linewidth]{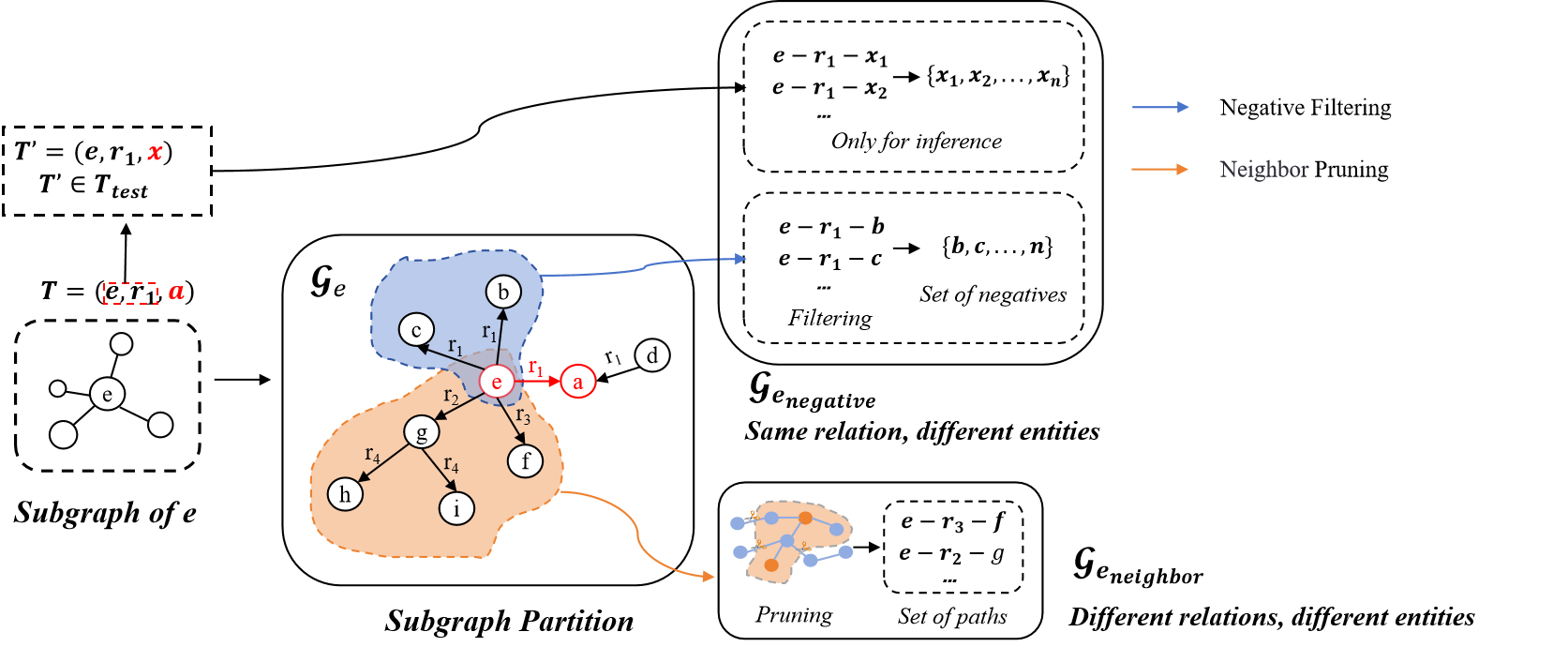} 
    \caption{Illustration of negatives and neighbors information for subgraph segmentation. Negatives represent other answers for $(e, r_1)$ within the training set, neighbors represent triples connected to $e$ in the training set where the relationship is not $r_1$, and $x$ represents other answers for $(e, r_1)$ in the test set. }
    \label{fig:subgraph}
\end{figure}

To maximize the effective information for reasoning, we propose an algorithm that extracts subgraphs centered on entities within the KG. Specifically, as shown in Figure \ref{fig:subgraph}, we obtain the subgraph $\mathcal{G}_e$ of the known entity $e$, which contains all the associated information of entity $e$. For example, if the triple we currently need to predict is $T = (e, r_1, a)$, where $a$ is the entity to be predicted, we use subgraph partitioning to obtain more fine-grained set of negatives and set of neighbors, denoted as $\mathcal{G}_{e_{\text{negatives}}}$ and $\mathcal{G}_{e_{\text{neighbors}}}$. $\mathcal{G}_{e_{\text{negatives}}}$ primarily consist of different answers to the same incomplete triple, and during inference, we use more negatives from the test set to encourage LLM to generate more potential answers. $\mathcal{G}_{e_{\text{neighbors}}}$ represent neighbors information centered around $e$, providing context to assist the LLM's inference. $\mathcal{G}_{e_{\text{negatives}}}$ and $\mathcal{G}_{e_{\text{neighbors}}}$ are two independent parts that are mapped to different modules of the problem during model training.

\subsubsection{Filtering negatives}\label{sec:filter}
It is crucial to prevent information leakage between the training set and the test set. The test set must contain data that the model has never seen to fairly evaluate its generalization ability. We describe how to generate negative samples for positive samples during training to enhance the learning effect of LLMs while preventing information leakage.

When generating subgraph for the incomplete triple $(e, r_i)$, the subgraph $\mathcal{G}_e$ is constructed with $e$. This subgraph $\mathcal{G}_e$ includes all entities and relations directly connected to $e$.
\begin{equation}
\mathcal{G}_e = \{(e, r, x) \in \mathcal{T}_{\text{train}} \mid r \in \mathcal{R}, e, x \in \mathcal{E}\} 
\end{equation}
From $\mathcal{G}_e$, we then extract the set of triple $\mathcal{T}_{e,r_i}$ that include $e$ and have the relationship $r_i$.
\begin{equation}
\mathcal{T}_{e,r_i} = \{(e, r_i, x) \in \mathcal{G}_e\}
\end{equation}
Finally, we let $\mathcal{T}_{true}$ be the triple containing the correct entity, then remove $\mathcal{T}_{true}$ from $\mathcal{T}_{e,r_i}$ to obtain the set of all negative samples, denoted as $\mathcal{T}^{-}_{e,r_i}$. 
\begin{equation}
\mathcal{T}^{-}_{e,r_i} = \mathcal{T}_{e,r_i} \setminus \{\mathcal{T}_{true}\}
\end{equation}
These negatives can prevent the LLM from generating existing facts in KG and encourage the generation of more new answers.

\subsubsection{Pruning neighbors}\label{sec:pruning}
We extract neighbors directly related to $e$ from $\mathcal{G}_e$. To improve efficiency and accuracy, we performed necessary pruning and compression on $\mathcal{G}_{e_{\text{neighbors}}}$.

Firstly, the input to the LLMs is usually subject to limitations, so we need to control the scale of all $\mathcal{G}_{e_{\text{neighbors}}}$ corresponding to the triples.  Specifically, if the shortest path $d(e, x)$ from entity $e$ to entity $t/h$ is greater than 5\cite{zhang2024start}, those entities are pruned.
\begin{equation}
\mathcal{G}_{e} = \{(h, r, t) \in \mathcal{G} | d(e, h) \leq 5 \lor d(e, t) \leq 5\}
\end{equation}
Secondly, to ensure that $\mathcal{G}_{e_{\text{neighbors}}}$ does not contain negatives, we remove the content in $\mathcal{G}_{e_{\text{neighbors}}}$ from $\mathcal{G}_e$ and generate $\mathcal{G}_{e_{\text{neighbors}}}$ through set difference operations.
\begin{equation}
\mathcal{G}_{e_{\text{neighbors}}} = \mathcal{G}_e - \mathcal{G}_{e_{\text{negatives}}}
\end{equation}
When dealing with large graphs $\mathcal{G}$ even after pruning, the connected subgraphs generated for each entity may still be too large for LLMs. Therefore, we introduce $p$ to limit the path depth in the subgraph. The setting of $p$ directly affects the path length starting from entity $e$, thereby determining the size of the subgraph. We define $\mathcal{C}(e, p)$ as the contextual neighbors set centered on entity $e$ with a depth of $p$.
\begin{equation}
\mathcal{C}(e, p) = 
\begin{cases} 
\emptyset & \text{if } p = 0 \\
\{(e, r, e_1) \in \mathcal{G}_{e_{\text{neighbors}}}\} & \text{if } p = 1 \\
\{(e, r, e_1,...,e_p,r_p,e_\text{p+1}) \in \mathcal{G}_{e_{\text{neighbors}}}\} & \text{if } 1 < p \leq 5 
\end{cases}
\end{equation}

When $p=0$, the set of neighbors is empty; when $p=1$, only the first-order neighbors directly connected to $e$ are considered. 

When $p>1$, we extend to deeper paths. For example, when $p=2$, paths in the form of $(e, r_1, e_1, r_2, e_2)$ are considered. In summary, our goal is to accurately obtain the set $\mathcal{C}(e, p)$ of $e$ based on the hyperparameter $p$. 

\subsubsection{Information Merge}
Through Section \ref{sec:filter} and Section \ref{sec:pruning}, we obtained the set of negatives $\mathcal{T}^{-}_{e,r_i}$ and the set of neighbors $\mathcal{C}(e, p)$ from the incomplete triple.

We set the total number of required negatives and neighbors to $M$. For each triple's corresponding $\mathcal{T}^{-}_{e,r_i}$ and $\mathcal{C}(e, p)$, we determine the final information set according to the following rule: If $|\mathcal{T}^{-}_{e,r_i}| \geq M$, we will randomly sample $M$ elements from $\mathcal{T}^{-}_{e,r_i}$ to ensure that the size of the final information set does not exceed $M$. If $|\mathcal{T}^{-}_{e,r}| < M$, to reach the set size, we will randomly sample from $\mathcal{C}(e, p)$ to fill up to $M$ elements. Specifically, we need to sample $M - |\mathcal{T}^{-}_{e,r_i}|$ number of elements from $\mathcal{C}(e, p)$ as shown in:
\begin{equation}
\mathcal{D}(e, M) = 
\begin{cases} 
\text{RandomSample}(\mathcal{T}^{-}_{e,r}, \min(|\mathcal{T}^{-}_{e,r}|, M)) & \text{if } |\mathcal{T}^{-}_{e,r}| \geq M \\
\mathcal{T}^{-}_{e,r} \cup \text{RandomSample}(\mathcal{C}, \min(|\mathcal{C}|, M - |\mathcal{T}^{-}_{e,r}|)) & \text{if } |\mathcal{T}^{-}_{e,r}| < M 
\end{cases}
\end{equation}
The final information set $\mathcal{D}(e, M)$ consists of the adjusted $\mathcal{T}^{-}_{e,r_i}$ and the supplemented $\mathcal{C}$. It should be noted that if $|\mathcal{T}^{-}_{e,r}| \geq M$, then the supplemented $\mathcal{C}$ will be empty. This method ensures that the number of information sets exactly meets the predetermined needs while maintaining the diversity of the dataset and the balance of information.

\subsection{QA Template Mapping}
LLMs have demonstrated outstanding capabilities in NLP tasks. Based on this, we employ a simple QA format to implement the link prediction task. For a missing triple $(h, r, ?)$, we designed a basic QA template $Prompt_{basic}$:

\begin{equation}
Prompt_{basic} = \text{Please complete this triple: (h, r, ?). h means $Desc_h$}
\end{equation}

Here, $Desc_h$ represents the textual description of entity $h$, using semantically enhanced descriptions from CP-KGC\cite{yang2024enhancing}. Adding $Desc_h$ helps the LLM understand the entity's meaning and effectively distinguishes the polysemy of the entity in different contexts. For each triple in $\mathcal{G}$, subgraphs are obtained through subgraph partitioning, resulting in $\mathcal{G}_{\text{part}} = \{\mathcal{G}_1, \mathcal{G}_2, \ldots, \mathcal{G}_m\}$. For any pair $(e, r)$, the corresponding subgraph is denoted as $\mathcal{G}_{e,r}$, either $\mathcal{T}^{-}_{e,r}$ or $\mathcal{C}(e, p)$ may be empty:
\begin{equation}
\mathcal{G}_{e,r} = \{\mathcal{T}^{-}_{e,r}, \mathcal{C}(e, p)\}
\end{equation}

When $\mathcal{T}^{-}_{e,r}$ is not empty, $e$ and $r$ in $\mathcal{T}^{-}_{e,r}$ are the same. $\mathcal{T}^{-}_{e,r}$ can be represented as $[e'_1, e'_2, \ldots, e'_n]$. At this point, $Prompt_{Negative}$ is:
\begin{equation}
Prompt_{Negative} = \text{Please give an answer outside the list: [$e'_1$, $e'_2$, \ldots, $e'_n$]}
\end{equation}
When $\mathcal{C}(e, p)$ is not empty, $\mathcal{C}$ includes triples or paths related to $e$. This neighbors provide LLMs with more relevant knowledge for reasoning. At this point, $Prompt_{Neighbors}$ is:
\begin{equation}
Prompt_{Neighbors} = \text{The neighbors of $e$ are as follows: }\mathcal{C}(e, p)
\end{equation}
Finally, the questions formed by integrating this information are input into the LLM for training:
\begin{equation}
Prompt_{e,r} = Prompt_{basic} + Prompt_{Negative} + Prompt_{Neighbors}
\end{equation}

\subsection{Model Training}
Instruction-based fine-tuning is a training strategy for LLMs designed to optimize their performance on specific tasks by providing instructions and examples. While full-parameter fine-tuning is effective, it encounters significant challenges with large-scale data.

LoRA is an efficient fine-tuning technique that quickly adapts to tasks by adjusting a small number of model parameters, reducing computational resource demands while maintaining the model's generalization ability. During LoRA fine-tuning, the pre-trained weight matrix $\mathbf{W} \in \mathbb{R}^{d \times k}$ is updated by the product of two low-rank matrices, $\mathbf{W}_A \in \mathbb{R}^{d \times r}$, $\mathbf{W}_B \in \mathbb{R}^{r \times k}$,  and the rank $r \ll min(d,k)$. Consequently, the updated weights can be represented as:
\begin{equation}
\mathbf{W}' = \mathbf{W} + \mathbf{W}_A \mathbf{W}_B
\end{equation}
Given a pre-trained LLM denoted as $\mathcal{M}$, with parameter count $\theta$, the training set includes $2 \times |\mathcal{T}_{\text{train}}| - 1$ instances $\{\text{Question}_e, \text{Answer}_e\}$. The purpose of fine-tuning is to minimize the following loss function:
\begin{equation}
\theta^* = \arg \min_{\theta'} \sum_{i=0}^{2 \times |\mathcal{T}_{\text{train}}| - 1} \mathcal{L}(\mathcal{M}(Q| \theta'), A)
\end{equation}
Where $\mathcal{M}(\theta')$ is the output of the fine-tuned LLM with parameter $\theta'$, $Q$ is the question, and $A$ is the answer.

\subsection{Evaluation of Generative KGC}\label{sec:evaluation}
The main difference between generative KGC and traditional KGC is that generative KGC directly produces the answer it considers most probable. In the real world, many questions do not have a single correct answer. LLMs are trained on vast text datasets, enabling them to learn language nuances and a broad range of knowledge, thereby possessing the powerful ability to generate and understand human language. Therefore, even if the answer given by an LLM is not in the standard answers of the test set, it may still be correct, as there can be multiple reasonable answers to the same question.

\begin{figure}[h!]
    \centering
    \includegraphics[width=0.75\linewidth]{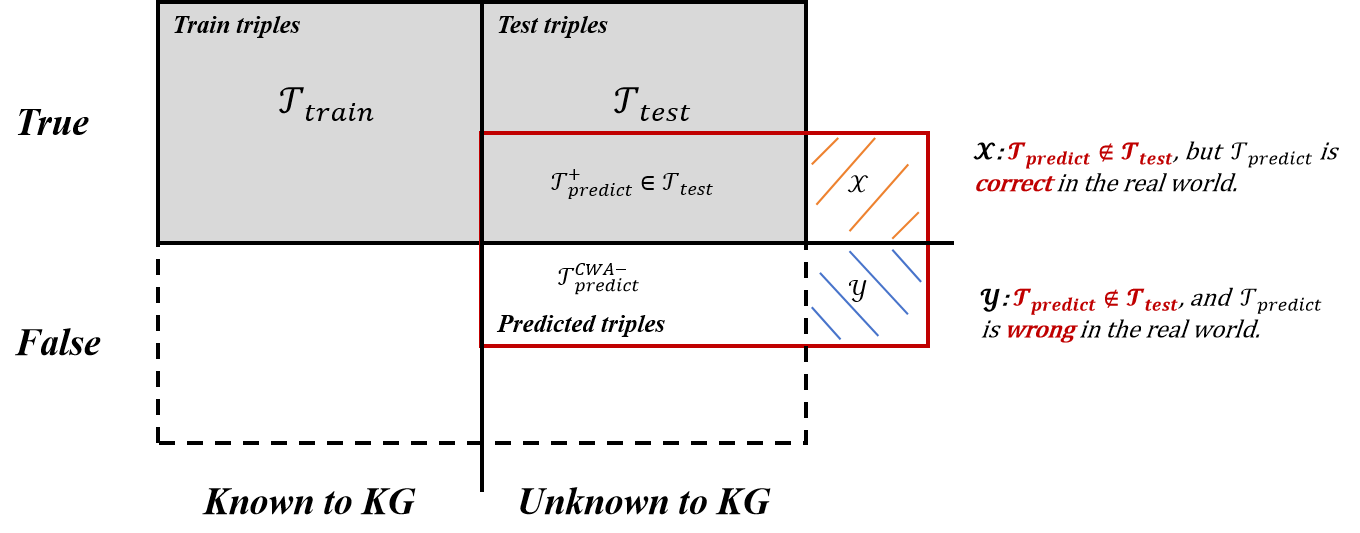} 
    \caption{The real completion result of LLM in the CWA.}
    \label{fig:method-GSKGC}
\end{figure}

As illustrated in Figure \ref{fig:method-GSKGC}, excluding parts $\mathcal{X}$ and $\mathcal{Y}$, the closed-world assumption (CWA) implies that triples not present in the KG are considered false. Under the CWA, triples in the training set are known to be true, while test triples are of unknown truth value. The predicted set of triples, $\mathcal{T}_{\text{predict}}$, can be divided into two subsets: $\mathcal{T}^{+}_{\text{predict}}$, containing true triples from the test set, and $\mathcal{T}^{\text{CWA-}}_{\text{predict}}$, containing false triples not in $\mathcal{T}_{\text{test}}$. We can achieve more accurate outcomes, thus bridging the gap between CWA-based and open-world assumption (OWA)-based KGC.
\begin{equation}
\mathcal{T}_{\text{predict}} = \mathcal{T}^{+}_{\text{predict}} + \mathcal{T}^{-}_{\text{predict}} \tag{15}
\end{equation}
\begin{equation}
\mathcal{T}^{-}_{\text{predict}} = \mathcal{T}^{\text{CWA-}}_{\text{predict}} + \mathcal{X} + \mathcal{Y} \tag{16}
\end{equation}
\begin{equation}
\mathcal{T}^{\text{LLM+}}_{\text{predict}} = \mathcal{T}_{\text{predict}} - \mathcal{T}^{\text{CWA-}}_{\text{predict}} - \mathcal{Y} \tag{17}
\end{equation}
We reevaluate the generative KGC results produced by LLMs from the CWA perspective, utilizing $\mathcal{X}$ and $\mathcal{Y}$. $\mathcal{T}^{\text{LLM+}}_{\text{predict}}$ signifies the true positive predictions by the LLM, and the final result should be the sum of $\mathcal{T}^{+}_{\text{predict}}$ and $\mathcal{X}$. For a detailed analysis, refer to Section\ref{sec:Advantages of GKGC}.

\section{Experiment}\label{sec:Experiment}
\subsection{Datasets}

\begin{table}[h!]
\centering
\caption{Statistical information of datasets.}
\label{tab:data}
\begin{tabular}{cccccc}
\hline
\textbf{Dataset} & \textbf{\#entity} & \textbf{\#relation} & \textbf{\#train} & \textbf{\#value} & \textbf{\#test} \\ \hline
WN18RR & 40943 & 11 & 86835 & 3034 & 3134 \\
FB15k-237 & 14541 & 237 & 272115 & 17535 & 20466 \\
FB15k-237N & 13104 & 93 & 87282 & 1827 & 1828 \\
ICEWS14 & 6869 & 230 & 74845 & 8514 & 7371 \\
ICEWS05-15 & 10094 & 251 & 368868 & 46302 & 46159 \\ \hline
\end{tabular}
\end{table}

We conducted experiments on several popular KGs, including three SKGs and two TKGs. The SKGs are WN18RR\cite{dettmers2018convolutional}, FB15k-237\cite{toutanova2015observed}, and FB15k-237N\cite{lv2022pre}. WN18RR is a subset of WordNet\cite{miller1995wordnet} that consists of English phrases and their semantic relations. FB15k-237 and FB15k-237N are derived from Freebase\cite{bollacker2008freebase} and contains numerous real-world facts. The TKGs include the widely used ICEWS14\cite{garcia2018learning} and ICEWS05-15\cite{li2021temporal} datasets. The statistics of these datasets are summarized in Table \ref{tab:data}. Each dataset is divided into training, validation, and test sets. 

\subsection{Baselines and Experiment Setup}
In our SKGC experiments, we used text-based KGC models such as STAR\cite{wang2021structure}, KG-BERT\cite{yao2019kg}, KG-S2S\cite{chen2022knowledge}, FTL-LM\cite{lin2023fusing}, and SimKGC\cite{wang2022simkgc} for comparison. For the LLM-based experimental setup, we selected KG-S2S-CD\cite{li2024contextualization}, SimKGC+MPIKGC-S\cite{xu-etal-2024-multi}, and SimKGC+CP-KGC\cite{yang2024enhancing}. For TKGC, we compared traditional models like TiRGN\cite{li2022tirgn}, HGLS\cite{zhang2023learning}, GenTKG\cite{liao2024gentkg}, TECHS\cite{lin2023techs}, and GPT-NeoX-20B-ICL\cite{black-etal-2022-gpt} with LLM-based models Llama2-7b-ICL\cite{touvron2023llama} and Llama2-7b-CoH\cite{luo2024chain}.

The foundational LLM models used were llama3-8b-instruct\cite{dubey2024llama} for SKGC and glm4-9b-chat\cite{glm2024chatglm} for TKGC. Specific fine-tuning parameters were: LoRA rank set to 8, LoRA alpha set to 16, LoRA dropout rate at 0.05, learning rate at 0.0001, and a training period of 1.0 epoch. The training equipment included 4 A-800-80GB PCIe graphics cards. In addition, we set $p=1$,$M=100$. In inference to obtain multiple results, the parameters are set as follows: enable sampling mode, set top-p to 0.95, top-k to 20, and use num return sequences to control the number of outputs generated by the model.

\subsection{Main Results}
We compared the performance of GS-KGC with baseline models. Experimental results on FB15k-237N, FB15k-237, and WN18RR are presented in Table \ref{tab:skgc}. Results on ICEWS14 and ICEWS05-15 datasets are shown in Table \ref{tab:tkgc}. From these comparisons, we made the following three key observations:

\begin{table}[htbp]
\centering
\caption{The performance comparison of different models in SKG. The best results are bolded, and the second-best results are underlined.}
\label{tab:skgc}
\resizebox{\textwidth}{!}{
\begin{tabular}{lcccccc}
\toprule
\textbf{Models} & \multicolumn{2}{c}{\textbf{FB15k-237N}} & \multicolumn{2}{c}{\textbf{FB15k-237}} & \multicolumn{2}{c}{\textbf{WN18RR}} \\
 & \textbf{Hits@1} & \textbf{Hits@3} & \textbf{Hits@1} & \textbf{Hits@3} & \textbf{Hits@1} & \textbf{Hits@3} \\
\midrule
\textbf{Text-based} & & & & & & \\
\hline
STAR\cite{wang2021structure}& - & - & 25.7 & 37.3 & 45.9 & 59.4 \\
KG-BERT\cite{yao2019kg} & - & - & - & - & 4.1 & 30.2 \\
FTL-LM\cite{lin2023fusing} & - & - & 25.3 & 38.6 & 45.2 & 63.7 \\
KG-S2S\cite{chen2022knowledge} & 28.2 & 38.5 & \underline{26.6} & \underline{40.4} & 53.1 & 59.5 \\
SimKGC\cite{wang2022simkgc} & 28.9 & 40.2 & 24.9 & 36.2 & \textbf{\underline{58.7}} & \textbf{\underline{71.7}} \\
\hline
\textbf{LLM-based} & & & & & & \\
\hline
KG-S2S-CD\cite{li2024contextualization} & 28.9 & 39.4 & - & - & 52.6 & 60.7 \\
MPIKGC-S\cite{xu-etal-2024-multi} & - & - & 24.5 & 36.3 & 52.8 & 66.8 \\
CP-KGC\cite{yang2024enhancing} & \underline{29.2} & \underline{40.3} & 26.0 & 37.5 & 58.0 & 68.3 \\
\hline
$\text{GS-KGC}_{\text{zero-shot}}$ & 4.4 & 6.9 & 3.4 & 5.5 & 0.9 & 2.4 \\
$\text{GS-KGC}_{\text{qa}}$ & 23.6 & 41.4 & 16.3 & 35.6 & 28.7 & 49.1 \\
GS-KGC & \textbf{34.0} & \textbf{45.9} & \textbf{28.0} & \textbf{42.6} & 34.6 & 51.6 \\
\bottomrule
\end{tabular}
}
\end{table}

\textbf{Results in  $\text{GS-KGC}_{\text{zero-shot}}$, $\text{GS-KGC}_{\text{qa}}$ and GS-KGC}: In Tables \ref{tab:skgc} and \ref{tab:tkgc}, $\text{GS-KGC}_{\text{zero-shot}}$ represents the model's output results without fine-tuning.  $\text{GS-KGC}_{\text{qa}}$ refers to using only the fine-tuned model, while GS-KGC employs negatives and neighbors. From the result in $\text{GS-KGC}_{\text{zero-shot}}$, it can be seen that the model is almost unable to generate the predicted entities without fine-tuning. In SKGC, GS-KGC outperformed $\text{GS-KGC}_{\text{qa}}$ in Hits@1 by 10.4\%, 11.7\%, and 5.9\% on the FB15k-237N, FB15k-237, and WN18RR, respectively. However, the improvements in Hits@3 were more modest, with an average increase of only 4.8\% across the three datasets. A similar pattern was seen in the TKGC, where GS-KGC achieved 8\% and 3.6\% improvements over $\text{GS-KGC}_{\text{qa}}$ in Hits@1 for the ICEWS14 and ICEWS05-15, respectively, while the average increase in Hits@3 was 5.4\%. In the generative link prediction task, when a single answer is required, the model produces stable outputs by disabling sampling and setting top-k to 1. However, in the case of one-to-many questions, the model generates the answer it finds most probable. Negatives and neighbors help the model refine its predictions in these cases. When multiple answers are needed, such as for Hits@3 or Hits@10, the model uses repeated sampling with appropriate top-p and top-k settings to generate a wider range of plausible answers. This approach enables $\text{GS-KGC}_{\text{qa}}$ to perform better on Hits@3. Overall, negatives and neighbors are critical in improving the accuracy of LLMs in generative link prediction tasks.

\textbf{Results in SKGC}: In SKGC, GS-KGC outperforms text-based KGC methods and their LLM-enhanced variants on the FB15k-237N, and FB15k-237 but shows slightly worse performance on WN18RR. Specifically, GS-KGC achieved improvements of 4.8\% and 5.6\% over CP-KGC on Hits@1 and Hits@3, respectively, on the FB15k-237N. Additionally, GS-KGC outperformed KG-S2S on the FB15k-237, with increases of 4.8\% and 4.5\% on Hits@1 and Hits@3, respectively. The poorer performance on WN18RR can be attributed to the unique characteristics of its entities, where the same word often has multiple meanings. For instance, in stool NN 2, "stool" is the entity, "NN" indicates that it is a noun, and "2" represents its second meaning in WN18RR. LLMs struggle to directly understand these complex meanings. While they can utilize known information during reasoning, they are prone to hallucinations, generating entities that do not belong to the WN18RR dataset. This highlights the model's limitations in handling highly polysemous entities, indicating a need for further optimization to reduce errors and non-target entity generation. In the forward and backward test sets, the LLM generated 1,220 and 1,424 entities, respectively, that do not exist in WN18RR. These account for 38.9\% and 45.3\%, respectively. Entities like "stool NN 2" led to additional results that were all incorrect.

\begin{table}
\centering
\caption{The performance comparison of different models in TKG. The best results are bolded, and the second-best results are underlined.}
\label{tab:tkgc}
\resizebox{\textwidth}{!}{
\begin{tabular}{lcccc}
\toprule
\textbf{Models} & \multicolumn{2}{c}{\textbf{ICEWS14}} & \multicolumn{2}{c}{\textbf{ICEWS05-15}} \\
 & \textbf{Hits@1} & \textbf{Hits@3} & \textbf{Hits@1} & \textbf{Hits@3} \\
\midrule
TiRGN\cite{li2022tirgn} & 32.8 & 48.1 & 37.9 & 54.4 \\
HGLS\cite{zhang2023learning} & \underline{36.8} & 49.0 & 36.0 & 52.5 \\
GenTKG\cite{liao2024gentkg} & 36.5 & 48.8 & 37.8 & 54.1 \\
GPT-NeoX-20B-ICL\cite{black-etal-2022-gpt} & 29.5 & 40.6 & 36.7 & 50.3 \\
TECHS\cite{lin2023techs} & 34.6 & 49.4 & 38.3 & \underline{54.7} \\
LLaMA2-7b-ICL\cite{touvron2023llama} & 28.6 & 39.7 & 35.3 & 49.0 \\
LLaMA2-7b-COH\cite{luo2024chain} & 34.9 & 47.0 & \underline{38.6} & 54.1 \\
\hline
$\text{GS-KGC}_{\text{zero-shot}}$ & 13.0 & 24.2 & 7.6 & 13.1\\
$\text{GS-KGC}_{\text{qa}}$ & 34.7 & \underline{50.6} & 35.1 & \underline{54.7} \\
GS-KGC & \textbf{42.7} & \textbf{58.7} & \textbf{38.7} & \textbf{57.4} \\
\bottomrule
\end{tabular}
}
\end{table}

\textbf{Results in TKGC}: GS-KGC shows strong performance in TKG, particularly on the ICEWS14 and ICEWS05-15, outperforming comparable models. For the Hits@1, GS-KGC surpasses HGLS by 5.9\% on ICEWS14, while on ICEWS05-15, its results are similar to those of other models. In the Hits@3 , GS-KGC outperforms TECHS by 9.3\% on ICEWS14 and by 2.7\% on ICEWS05-15. Furthermore, $\text{GS-KGC}_{\text{qa}}$ performs better than other baseline models on ICEWS14 and remains competitive on ICEWS05-15, where it matches the baseline performance. Unlike SKG, TKG is to predict future entities based on historical observations. In this context, LLMs can effectively learn and adapt to these tasks through LoRA. Since the data relationships in TKGC are relatively simpler and more straightforward than in SKGC, using the QA-only mode can achieve good results.

These observations verify the effectiveness of GS-KGC. GS-KGC shows great potential in integrating negatives and neighbors in LLMs for KGC. This feature allows GS-KGC to flexibly and continuously learn KGC from new entities and relations in a KG. Advances in LLMs can potentially further enhance GS-KGC's accuracy.

\section{Analysis}\label{sec:Analysis}
\subsection{Ablation Studies}

We conducted ablation studies on the FB15k-237N, ICEWS14, and WN18RR to assess the impact of negatives and neighbors on KGC. The results, shown in Table \ref{tab:ablation}, define \textit{w/o neighbors} as the ablation without neighbors, \textit{w/o negatives} as the ablation without negatives, and \textit{w/o neighbors+negatives} as the ablation without both neighbors and negatives. This means that the only variable is whether contextual information is included.  The table reveals that \textit{w/o neighbors} performs better than \textit{w/o negatives}. For example, in the FB15k-237N, Hits@1 and Hits@3 drop by only 0.2\% and 2.3\%, respectively, in the \textit{w/o neighbors} experiment. In contrast, the \textit{w/o negatives} experiment results in sharper declines of 10.7\% and 5.6\%. Furthermore, the \textit{w/o negatives} setting underperforms compared to \textit{w/o neighbors+negatives.}

\begin{table}[htbp]
\centering
\caption{The ablation results of GS-KGC.}
\label{tab:ablation}
\resizebox{\textwidth}{!}{
\begin{tabular}{lcccccc}
\hline
\textbf{Ablation}       & \multicolumn{2}{c}{\textbf{FB15k-237N}} & \multicolumn{2}{c}{\textbf{ICEWS14}} & \multicolumn{2}{c}{\textbf{WN18RR}} \\
                        & \textbf{Hits@1} & \textbf{Hits@3} & \textbf{Hits@1} & \textbf{Hits@3} & \textbf{Hits@1} & \textbf{Hits@3} \\
\hline
GS-KGC                  & 34.0 & 45.9 & 42.3 & 58.7 & 34.6 & 51.6     \\
\hline
GS-KGC w/o neighbors    & 33.8 & 43.6 & 40.1  & 57.8 & 31.8 & 50.2    \\
$\Delta$                & -0.2 & -2.3 & -2.2 &-0.9 & -2.8  & -1.4  \\
\hline
GS-KGC w/o negatives     & 23.3 & 40.3 & 34.5 & 51.8 & 27.8 & 46.2  \\
$\Delta$                & -10.7 & -5.6  & -7.8 & -6.9 & -6.8  &-5.4   \\
\hline
GS-KGC w/o neighbors+negatives     & 23.6 & 41.4 & 34.7 & 50.6 & 28.7 & 49.1  \\
$\Delta$                & -10.4 & -4.5  & -7.6 & -8.1 & -5.9  &-2.5   \\
\hline
\end{tabular}
}
\end{table}

The significant drop in the \textit{w/o negatives} results, even below \textit{w/o neighbors+negatives}, suggests that neighbors alone does not improve model performance. However, the \textit{w/o neighbors} results show that combining negatives with neighbors can further enhance performance. We believe this is because, in the real world, an incomplete triple often corresponds to a question with multiple potential answers. Using known results as negatives can motivate the model to explore more possibilities, thereby enhancing its generalization capability. On the other hand, when only neighbors are added, the model gains more background information. However, this information highly overlaps with the background of the known answers, resulting in suboptimal performance in the test set. This is further illustrated in the results of GS-KGC.

\subsection{Hyperparameter Analysis}\label{sec:hyperparameter}
The reasoning ability of LLMs is often limited by the amount of available context. In Tables \ref{tab:skgc} and \ref{tab:tkgc}, we set the total number of negatives and neighbors for GS-KGC to 100. To balance model reasoning performance and resource consumption, we conducted parameter analysis experiments on three datasets—FB15k-237N, ICEWS14, and WN18RR—to explore the impact of different M values (0, 20, 40, 60, 80, 100) on model performance.

\begin{figure}[h!]
    \centering
    \includegraphics[width=1.0\linewidth]{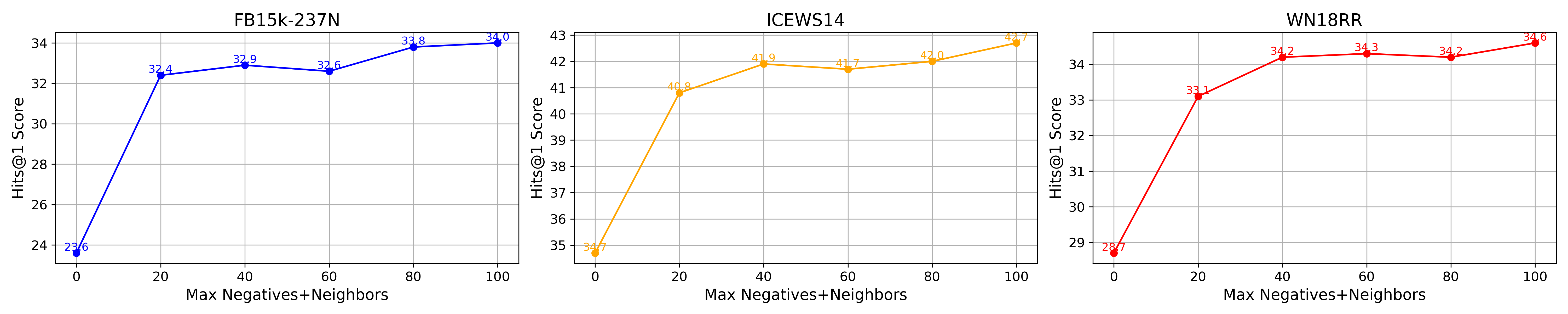} 
    \caption{The line chart comparison of Hits@1 for different datasets as parameter M varies.}
    \label{fig:hyperparameter}
\end{figure}

Figure \ref{fig:hyperparameter} shows the trend in GS-KGC’s Hits@1 performance. As the value of M increases from 0 to 20, the model's performance improves by 8.8\%, 6.1\%, and 4.4\% on the FB15k-237N, ICEWS14, and WN18RR, respectively, with an average gain of 6.4\%. When M exceeds 20 and approaches 100, the performance gains become less significant. Ablation studies reveal that negatives contribute more to GS-KGC's performance than neighbors, though both have complementary effects. Following the principle of prioritizing negatives (Equation 5), setting M to 20 enables negatives to help the model exclude obvious known answers, enhancing overall performance.

We further examined why Hits@1 improves significantly at M = 20. The one-to-many problem is widely present in the training and test sets of FB15k-237N, ICEWS14, and WN18RR.  Negatives help GS-KGC better distinguish these repeated patterns, explaining the performance boost at M = 20. Additionally, retrieving high-quality context from known datasets improves response accuracy. Optimizing context recall reduces both training time and resource use, making it advantageous for larger datasets.

\subsection{Advantages of GS-KGC}\label{sec:Advantages of GKGC}

In Section \ref{sec:evaluation}, we highlighted that generative KGC not only identifies new triple relations within a KG but also suggests entities outside the KG as potential answers. This capability aligns with the core goal of KGC, which goes beyond link prediction in closed datasets. For example, in the FB15k-237N, with 8,226 test triples, we evaluated cases where predictions failed using Qwen-plus and GPT-4o to judge the plausibility of the predicted entities. If both models deemed the prediction reasonable, we considered the triple potentially valid. This approach provides deeper insights into the predictions.

\begin{figure}[h!]
    \centering
    \includegraphics[width=1.0\linewidth]{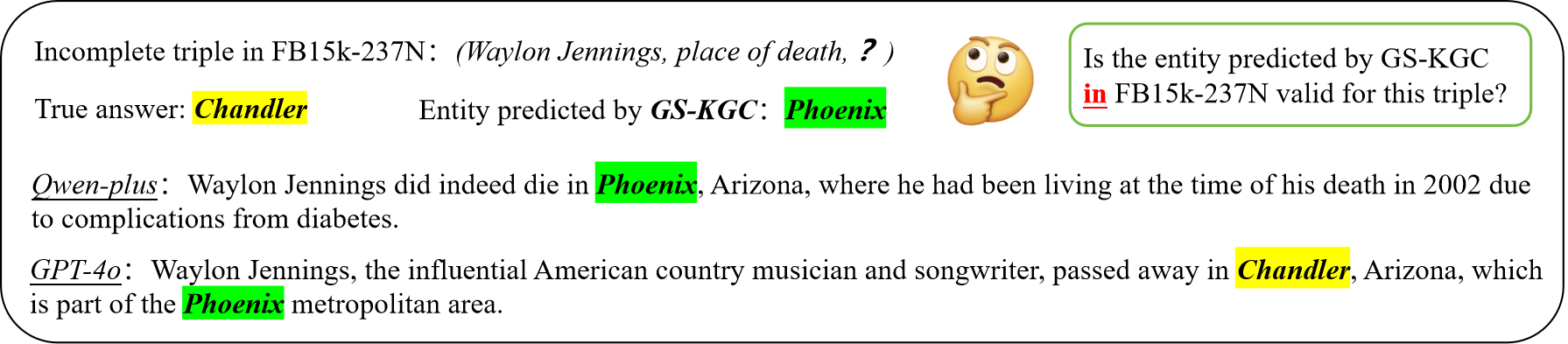} 
    \caption{GS-KGC predicted an entity that exists in FB15k-237N, resulting in a new triple.}
    \label{fig:in}
\end{figure}

\begin{figure}[h!]
    \centering
    \includegraphics[width=1.0\linewidth]{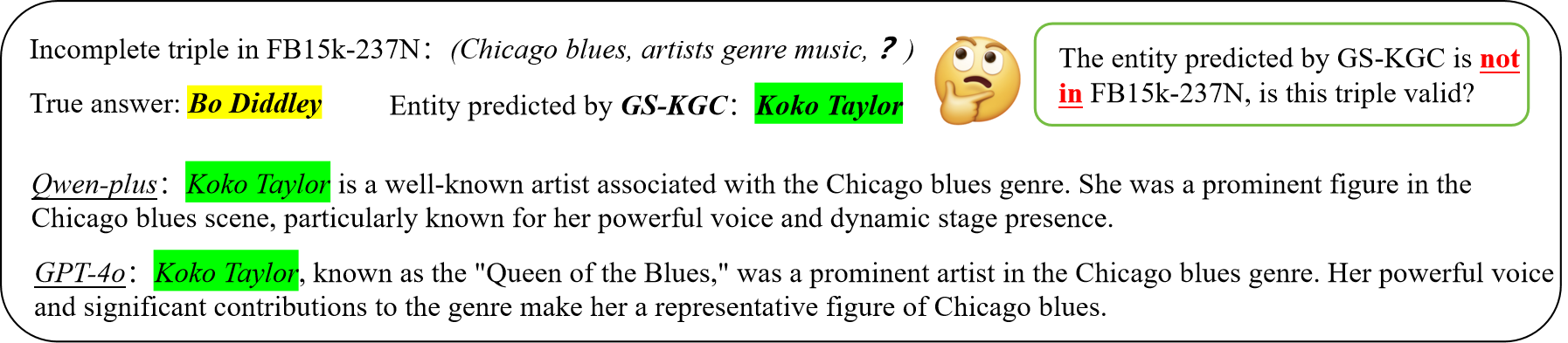} 
    \caption{GS-KGC predicted an entity that does not exist in FB15k-237N, revealing a new fact beyond the dataset. }
    \label{fig:not in}
\end{figure}

Two main scenarios in GS-KGC’s predictions merit attention. \textit{\textbf{A}}: The predicted entity exists in FB15k-237N, but the triple itself does not. For example, in Figure \ref{fig:in}, the triple \textit{(Waylon Jennings, place of death, ?)} has Chandler as the test set answer, while GS-KGC predicted Phoenix, part of the Phoenix metropolitan area, which includes Chandler. As both Qwen-plus and GPT-4o validated Phoenix as a reasonable prediction, we consider \textit{(Waylon Jennings, place of death, Phoenix)} plausible, even though it is absent from FB15k-237N. Thus, GS-KGC helps reveal new triple in KG. \textit{\textbf{B}}: The predicted entity does not exist in FB15k-237N. In Figure \ref{fig:not in}, the triple \textit{(Chicago blues, artists genre music, ?)} has \textit{Bo Diddley} as the test set answer, but GS-KGC predicted the \textit{Koko Taylor}. In this example, there are many famous artists in \textit{Chicago blues} music, including \textit{Koko Taylor} and \textit{Bo Diddley}. Therefore, it is a reasonable prediction for \textit{(Chicago blues, artists genre music, Koko Taylor)}.

GS-KGC not only aids in discovering new facts within KGs but also bridges the gap between closed-world and open-world KGC. This ability stems from the extensive data processed by LLMs during pre-training, which informs their predictions. Continuing to pre-train LLMs on domain-specific data, followed by fine-tuning on targeted datasets for KGC, could further enhance their capacity to uncover new facts. This approach broadens the practical utility and impact of KGC in specialized domains.

\subsection{How Model Size Affects Results}
In this section, we explore how the scale of LLMs affects the performance of KGC. We compared two model scales, Qwen2-1.5B-Instruct and Qwen2-7B-Instruct\cite{qwen2}. All other conditions were kept constant. The specific comparison results are shown in Table \ref{tab:size}.

\begin{table}[h!]
\centering
\caption{The Impact of Different Parameter Sizes of LLM on the Results.}
\label{tab:size}
\begin{tabular}{cccc}
\hline
Model size & FB15k-237N & ICEWS14 & WN18RR\\ \hline
\textit{GS–KGC$_{1.5B}$} & 24.2& 35.2& 16.3\\
\textit{GS–KGC$_{7B}$} & 32.7& 40.9& 32.5\\ \hline
\end{tabular}
\end{table}

On the FB15k-237N, ICEWS14, and WN18RR, the performance differences between the two models are 8.5\%, 5.7\%, and 16.2\%, respectively. In the 1.5B parameter model, we observed certain differences compared to traditional methods. This indicates that within a certain parameter range, increasing the model size can effectively enhance performance. A larger model implies stronger reasoning capabilities. With continuous advancements in LLM technology and iterative updates of open-source models, we anticipate that reasoning capabilities will continue to improve, paving the way for the widespread application of KGC.

\section{Conclusion and future work}\label{sec:Conclusion}
In this paper, we present a framework called GS-KGC for KGC using generative LLMs. Due to the common challenge of one-to-many, LLMs may struggle with precision in generating target entities. To address this, we introduce techniques for negatives and neighbors to improve model accuracy. Specifically, we propose a negative sampling approach based on known answers and incorporate neighbors to strengthen LLMs’ understanding of facts. We improve the model’s reasoning by retrieving relevant contexts from existing datasets. Finally, we reframe the link prediction task as a QA problem, facilitating instruction-based fine-tuning.

In summary, our model offers three key advantages: (1) Theoretical: Unlike traditional KGC methods that rank all candidate entities, GS-KGC directly generates the best answer and can manage multi-answer predictions through beam search. (2) Experimental: In tests on SKG and TKG, our method outperformed existing inductive and LLM-based models across four datasets. (3) Application Potential: Our approach demonstrates notable advantages in generative KGC, not only discovering unknown triple within datasets but also generating new facts beyond them. Future vertical domain pre-training may further expand these capabilities, and we plan to explore this method’s potential across various specific fields.

\bibliographystyle{unsrt} 
\bibliography{references}






\end{document}